\documentclass[journal, onecolumn]{IEEEtran}
\usepackage{cite}
\usepackage{amsmath,amssymb,amsfonts,pifont}
\usepackage{algorithmic}
\usepackage{graphicx}
\usepackage{textcomp}
\usepackage{multirow}
\usepackage{adjustbox}
\usepackage{enumitem}
\usepackage{lscape}
\usepackage[numbers]{natbib}
\usepackage{afterpage}
\usepackage{balance}
\usepackage{graphicx}
\usepackage{latexsym}
\usepackage{amssymb}
\usepackage{amsmath}
\usepackage{balance}
\usepackage[ruled,vlined]{algorithm2e}
\ifCLASSINFOpdf
\else
\fi
\begin{document}
%
\title{Federated Split Learning with Only Positive Labels for resource-constrained IoT environment}
%
%

\author{Praveen~Joshi,
        Chandra~Thapa,
        Mohammed~Hasanuzzaman,
        Ted~Scully,
        and~Haithem~Afli
\thanks{\hskip-8pt Praveen Joshi, Mohammed Hasanuzzaman, Haithem Afli and Ted Scully are with the Computer Science Department, Munster Technological University, Rossa Ave, Bishopstown, Cork, T12 P928, Ireland. 
}
\thanks{\hskip-8pt Chandra Thapa is with Data61, Commonwealth Scientific and Industrial Research Organization (CSIRO), Sydney, 2122, Australia. 
}
\thanks{\hskip-8pt Corresponding author: Praveen Joshi (e-mail: praveen.joshi@mycit.ie).}
}

\maketitle

\begin{abstract}
Distributed collaborative machine learning (DCML) is a promising method in the Internet of Things (IoT) domain for training deep learning models, as data is distributed across multiple devices. A key advantage of this approach is that it improves data privacy by removing the necessity for the centralized aggregation of raw data but also empowers IoT devices with low computational power. Among various techniques in a DCML framework, federated split learning, known as splitfed learning (SFL), is the most suitable for efficient training and testing when devices have limited computational capabilities. 

Nevertheless, when resource-constrained IoT devices have only positive labeled data, multiclass classification deep learning models in SFL fail to converge or provide suboptimal results.
To overcome these challenges, we propose \emph{splitfed learning with positive labels} (SFPL). SFPL applies a random shuffling function to the smashed data received from clients before supplying it to the server for model training. Additionally, SFPL incorporates the local batch normalization for the client-side model portion during the inference phase. Our results demonstrate that SFPL outperforms SFL: (i) by factors of 51.54 and 32.57 for ResNet-56 and ResNet-32, respectively, with the CIFAR-100 dataset, and (ii) by factors of 9.23 and 8.52 for ResNet-32 and ResNet-8, respectively, with CIFAR-10 dataset. Overall, this investigation underscores the efficacy of the proposed SFPL framework in DCML.
\end{abstract}


%
\IEEEpeerreviewmaketitle

\section{Introduction}
\label{sec:intro}
Federated split learning is a distributed collaborative machine learning (DCML) approach that combines federated learning (FL)  \cite{konevcny2016federated} and split learning (SL) \cite{joshi2022performance}. This way, it jointly leverages the benefits of FL and SL, such as parallel computations across devices for faster DCML and splitting the model into multiple portions to support resource-constrained IoT devices. Federated split learning is also known as splitfed learning (SFL)~\cite{thapa2022splitfed}.  
SFL partitions a deep learning model and assigns only a portion of the model to each device in DCML. The devices with data are clients, and the rest are servers; their model portion is called client-side and server-side models, respectively. The output of a client-side model is referred to as smashed data.

\begin{figure*}[t]
  \centering
  \includegraphics[width=6.5in]{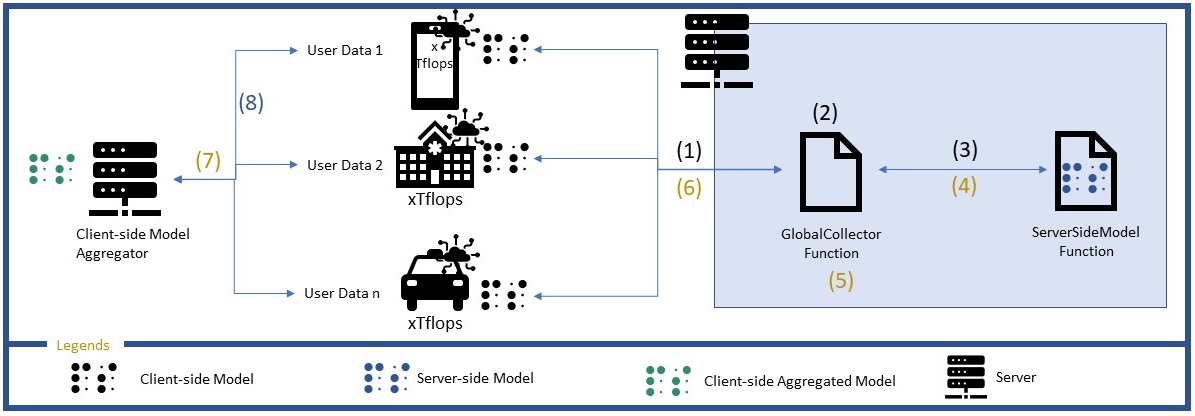}
  \caption{An overview of splitfed learning with positive labels (SFPL).}
  \label{fig:diagram}
\end{figure*}

\begin{figure}[t]
  \centering
  \includegraphics[width=3.2in]{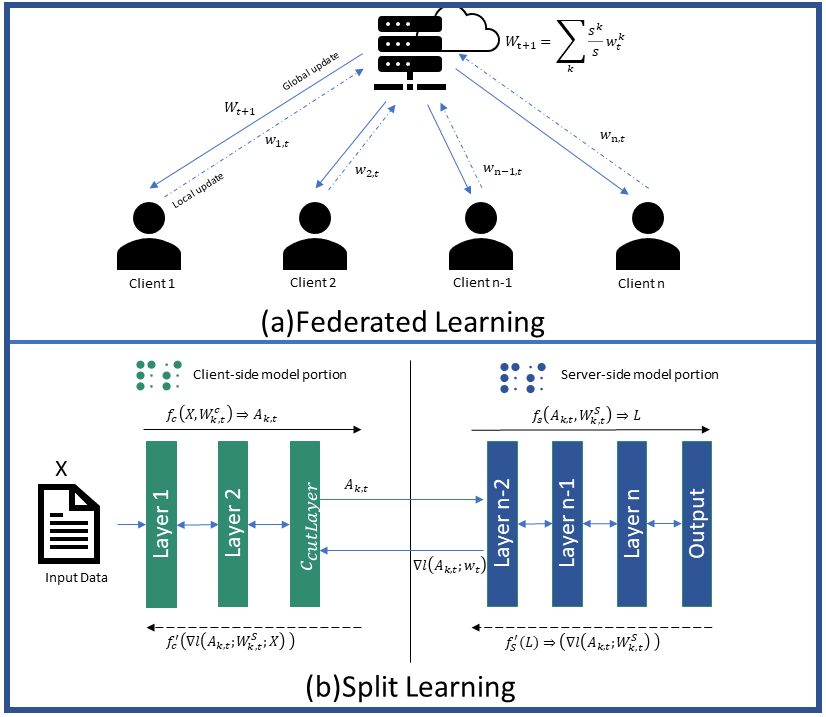}
  \caption{An overview of federated learning and split learning.}
  \label{fig:diagram2}
\end{figure}

Despite the benefits and potential of SFL, its applicability is only demonstrated in scenarios where the data source devices, called clients, either have independent and identical distribution (IID) data or imbalanced data distribution. However, in non-IID data distribution scenarios, SFL's performance is poor~\cite{thapa2022splitfed}. Moreover, in extreme non-IID assumptions, where the clients have only positive labels (each client has access to only one class), SFL exhibits a pronounced deficiency in learning~\cite{gao2021evaluation}. Extreme non-IID cases are expected in real-world scenarios. These scenarios encompass domains such as health care, wherein medical establishments are specialized in one category, \emph{e.g.}, radiology. Other examples include tracing food bio-origin, wherein the food is grown at one specific geographic location. Thus, more studies of SFL with only positive labels are entailed. In this regard, this is the first study to the best of our understanding.  
\par
This paper proposes splitfed learning with positive labels (SFPL). SFPL is tailored for DCML situations presenting only positive labels while maintaining equivalent performance when exposed to IID distributions. It incorporates a random shuffling function, which we call global collector function, in the server-side model training. As outlined in Figure \ref{fig:diagram}, the collector function is designed to accumulate a predetermined volume of smashed data prior to initiating the random shuffling process on the aggregated data, which is then directed to the server-side model training. This way, it creates smashed data stacks simulating an IID distribution for the server-side model training. However, for this task, the collector function alone is insufficient, so SFPL also attenuates the effect of the batch normalization layer by using local batch normalization for the client-side model portion to achieve the desired results during the inference phase. To assess the efficacy of SFPL, comprehensive experiments are conducted on commonly-used datasets, and the findings are compared with federated learning and SFL. Within the scope of this investigation, our attention is specifically directed toward the SFLv2 version of SFL. This version demonstrates a decreased memory footprint compared to its SFLv1 counterpart~\cite{thapa2022splitfed}, rendering it apt for deployment in resource-limited devices. The following are the highlights of our contributions:
\begin{itemize}
    \item \textbf{Splitfed learning with positive labels (SFPL)}: We propose a novel DCML to improve the deep learning model training for resource-constrained IoT clients in extreme non-IID data distributions.
    
    \item \textbf{Failure analysis of SFL}: As a motivation towards SFPL, we perform failure analysis of the SFLv2 under extreme non-IID data distribution scenarios, enabling the identification of its limitations, weaknesses, and potential areas of enhancement.
    
    \item \textbf{Performance analysis}: Empirical results consistently demonstrate the SFPL's superior performance (\emph{e.g.}, precision, recall) over SFLv2 across CIFAR-100 and CIFAR-10 datasets.    
\end{itemize}
\section{Related works}
To the best of our knowledge, this is the first research addressing the challenges associated with distributed learning, specifically when resource-limited IoT devices have exclusive access to positive labels. DCML with only positive labels has been performed in the context of FL~\cite{lin2022federated}. However, incorporating resource-constrained IoT devices introduces additional complexity to the problem if the client cannot run the full model. In this regard, SFL comes into the picture but is architecturally different from FL due to the SL components. As highlighted in research~\cite{gao2021evaluation}, all SFL variants fail to learn effectively when presented with a scenario where clients have access only to positive labels.\par
As mentioned in the introduction, various practical circumstances include clients having data with only positive levels. For instance, when tracing the bio-origin of food, the geographical location or country where the food is grown represents a positive label. In healthcare scenarios, the presence of specialization is often regarded as a positive label. Voice or face recognition models similarly consider the voice or face as positive labels. The proposed SFPL aids in safeguarding the privacy of user data amongst other participating users and servers, ensuring no direct data sharing. Nonetheless, the complexity of the data, whether it's grayscale images or multi-channel image datasets like RGB, can influence the degree of information leakage from activations sent to the client-side model. This leakage can be reduced in instances involving multi-channel datasets but could be greater with grayscale images~\cite{joshi2022performance}.
%

\section{Primer}
In this section, we aim to provide an overview of federated learning, split learning, and splitfed learning.
\subsection{Federated Learning}
The paradigm of federated learning (FL), presented in Figure~\ref{fig:diagram2} (a),  emphasizes the federated averaging (FedAvg) algorithm for local model aggregation~\cite{mcmahan2017communication}. Throughout the training phase, the server initially sets up the global model $W_t$, distributing it to all the participating clients. Upon receipt of the model $W_t$, every client $k$ proceeds to train the global model utilizing its local data, where $\mathcal{S}^k$ represents the number of training samples maintained by client $k$, and $\mathcal{S}$ signify the cumulative count of training samples across all clients. Subsequently, each client communicates the locally updated model $W_{k,t}$ back to the server. The server then undertakes the task of aggregating these models to refresh the global model to $W_{t+1}$. This iterative process, often called a 'round', persists until the model converges.
\subsection{Split Learning}
Distinct from FL, where every client trains the entire neural network, split learning (SL) ~\cite{vepakomma2018no} segments a neural network-based model into a minimum of two model portions. These model portions are then trained separately by the distributed parties, such as clients and servers. An elementary representation of SL is depicted in Figure~\ref{fig:diagram2} (b), with a cut layer acting as the division layer that splits the entire network into two model portions. The initial model portion $W^{\textup{C}}_{k,t}$ is trained and managed by the client, whereas the second model portion  $W^{\textup{S}}_{k,t}$ is under the server's control. During the training phase, both forward and backpropagation processes occur within the network. As shown in Figure~\ref{fig:diagram2} (b), the client initiates forward propagation on the input data and transmits the cut layer activations, referred to as the smashed data $\left(A_{k,t}\right)$, to the server. The server then conducts forward propagation on this smashed data, calculating the subsequent loss. The backpropagation begins from this point, yielding the gradients of the smashed data $\left(\nabla \ell\left(A_{k,t}; W^{\textup{S}}_{k,t}\right)\right)$. Once this gradient is computed, it is transmitted back to the client, initiating the client's backpropagation process. In the context of SL training/testing, the server does not have access to the clients' model portions and data, bolstering privacy. Along with privacy advantages, each client is only required to train a model portion that usually comprises a few layers, with the majority of layers housed in the server. Consequently, this reduces the computational load on the client. The learning performance (for instance, model accuracy and convergence) of SL has not yet been thoroughly explored when the data is non-Independent and Identically Distributed (non-IID) or unevenly distributed, which has been taken into consideration in this study.
\subsection{Splitfed Learning}
SL significantly diminishes the computational needs on the client side by operating only on a smaller model portion. Nonetheless, it necessitates sequential iterations over each client, thereby extending training time in instances where multiple clients are involved. In FL, clients typically parallelize their interactions with the server, thereby facilitating faster completion of training compared to SL. However, this method imposes a high computational overhead as each client must train the entire model. Recently, a hybrid approach combining the benefits of both SL and FL has been introduced, known as SFL~\cite{thapa2022splitfed}. Within SFL, all clients calculate independently and concurrently. These clients transmit/receive their smashed data to/from the server simultaneously. The synchronization of the client-side model portion, that is, the creation of the global client-side network, is achieved through the aggregation (for instance, weighted averaging) of all local networks on the client side on a separate server, termed the "fed server". Two distinct methodologies for server-side model portion synchronization were presented in~\cite{thapa2022splitfed}, leading to the creation of two specific versions of SFL:
\begin{itemize}
    \item SFLV1: The first methodology involves the independent and concurrent training of the smashed data of each client, resulting in an equal number of server-side model portions and clients. Subsequently, all the model portions are amalgamated (e.g., through weighted averaging) to form the global server-side network, a variant known as SFLV1.
    \item SFLV2: The second methodology entails sequential server-side model portion training on each client's smashed data (note that clients can still send their smashed data concurrently). This approach maintains a single copy of the model portion on the server side, which becomes the global server-side network once the main server has processed all the smashed data, referred to as SFLV2.
\end{itemize}

\section{Problem Setup}
\label{sec:ps}
\subsection{SFLv2 with positive labels}

In our setup, SFLv2 splits the full model into two portions; client-side model $W^\textup{C}$ and server-side model $W^\textup{S}$.
At the individual client level, a client-side model is denoted as $W^{\textup{C}}_k$, where $k \in [N], [N]:=\{1, \ldots, N\}$, $k$ represents the client, and $N$ indicates the maximum number of clients. 

We assume that each client has data from only one class, \emph{i.e.}, positive labels. Let $\mathcal{Z}$ be the specific set of instances, and $\mathcal{X}$ be a subset of $\mathcal{Z}$ employed in a specific batch. Given that there are $V$ classes indexed by a set $[V]$, the number of distinct sets $[V]$ equals $N$, with both $|[V]|$ and $N \in \mathbb{N}$. We assume that the $N$ clients collaboratively train for $t$ epochs, where $t \in \{1, \ldots, T\}$, and communicate with the server in a random order, once per epoch.
Assume that $g_{\boldsymbol{\theta}^\textup{C}}$ represents the function of client-side model parameters.
With this notation, the smashed data transmitted from the $k$-th client using its private data $\boldsymbol{X}_{k,t}$ can be formulated as:
\begin{equation}
\boldsymbol{A}_{k,t}= g_{\boldsymbol{\theta}^\textup{C}}(\boldsymbol{X}_{k,t}).
\end{equation}

Let $\mathcal{F} \subseteq\left\{f_s: \boldsymbol{A}_{k,t} \rightarrow \mathbb{R}^{V}\right\}$ be a set of scorer functions in the server-side model portion $W^{\textup{S}}$, which when given the smashed data $\boldsymbol{A}_{k,t}$, assigns a score to each of the $V$ classes. In particular, for $v \in[V], f_{s}(\boldsymbol{A}_{k,t})$ gives the probability of the $v$-th class for ${\boldsymbol{A}_{k,t}}$ received from the client-side model portion, as measured by the scorer $f_s \in \mathcal{F}$. For simplicity, scorers are of the form
\begin{equation}
f_s(\boldsymbol{A}_{k,t})= g_{\boldsymbol{\theta}^{\textup{S}}}({\boldsymbol{A}}_{k,t}),
\end{equation}
where $g_{\boldsymbol{\theta}^{\textup{S}}}:\boldsymbol{A}_{k,t} \rightarrow \mathbb{R}^v$ maps the instance $\boldsymbol{A}_{k,t}$ to a $v$-dimensional vector to produce the scores (or logits) for $V$ classes as $ g_{\boldsymbol{\theta}^{\textup{S}}}(\boldsymbol{A}_{k,t})$. 

In DCML with only positive labels, the $k$-th client has $n_k$ instances of all only label $k$. We postulate that each client possesses a subset of the comprehensive dataset, denoted as $\mathcal{S}=\cup_{k \in[{N}]} \mathcal{S}^k$. Here, the subset $\mathcal{S}^k$ represents the collection of $V=\sum_{k \in[N]} v_k$, where $v_k$ corresponds to the instances and label pairs jointly accessible to each client. This relationship can be explicitly defined as $k : \mathcal{S}^k=$ $\left\{\left(\boldsymbol{x}_{1}, k\right), \ldots,\left(\boldsymbol{x}_{n}, k\right)\right\} \subset \mathcal{X} \times[V]$.

 Our objective is to minimize the loss function $\ell: \mathbb{R}^V \times[V] \rightarrow \mathbb{R}$ to derive a score function $\mathcal{F}$ that accurately classifies instances. The loss function $\ell$ assesses the server-side scorer $f_s$ quality, considering $y^j_k$ as the corresponding label to the smashed data $\boldsymbol{A}^j_{k,t}$ corresponding to a data sample $\boldsymbol{x_j}$. It operates on the input-output pairs $(\boldsymbol{A}^j_{k,t}, y^j_k)$ to minimize the empirical risk estimation based on the activations received from the $k$-th client, as follows:
\begin{equation}
\hat{f}=\underset{f \in \mathcal{F}}{\operatorname{argmin}}  \hat{\mathcal{R}}\left(f_s ; S^k\right):=\frac{1}{n_k} \sum_{j \in\left[n_k\right]} \ell \left(f\left({\boldsymbol{A}^j_{k,t}}\right), y^j_k\right) .
\end{equation}
In SFLv2, each client's $W^{\textup{C}}_k$ trains with the server-side model at least once but in random order as follows:
\begin{enumerate}
    \item Initially, all clients receive the client-side model portion with parameters $\boldsymbol{\theta}^{\textup{C}}$ and $W^\textup{C}$ to all clients. At the same time, the server-side receives the server-side model portion with $\boldsymbol{\theta}^\textup{S}$ and $W^\textup{S}$.
    \item For each client $k \in[N]$, the $k$-th client transmits the smashed data $\boldsymbol{A}_{k,t}$ to the server-side model $W^\textup{S}$. Utilizing the empirical risk estimation, the client-side and server-side model components are updated accordingly: \par
    For client-side:
    \begin{align}
    \label{eq_c1}
    & \boldsymbol{\theta}_{k,t}^\textup{C}=\boldsymbol{\theta}_t^\textup{C}-\eta \cdot \nabla_{\boldsymbol{\theta}^\textup{C}} \hat{\mathcal{R}}\left(g_{\boldsymbol{\theta}^\textup{C}}; \mathcal{S}^k\right) .
    \end{align}
    \begin{align}
    & W_{k,t}^\textup{C}=W_t^\textup{C}-\eta \cdot \nabla_{W^\textup{C}} 
    \label{eq_c2}
    \hat{\mathcal{R}}\left(g_{\boldsymbol{\theta}^\textup{C}}; \mathcal{S}^k\right) .
    \end{align}
    For server-side:
    \begin{align}
    & \boldsymbol{\theta}_{k,t}^\textup{S}=\boldsymbol{\theta}_t^\textup{S}-\eta \cdot \nabla_{\boldsymbol{\theta}^\textup{S}} \hat{\mathcal{R}}\left(f_{s} ; \boldsymbol{A}_{k,t}\right) .
    \end{align}
    \begin{align}
    & W_{k,t}^\textup{S}=W_{t}^\textup{S}-\eta \cdot \nabla_{W^\textup{S}} \hat{\mathcal{R}}\left(f_{\textup{s}} ;\boldsymbol{A}_{k,t}\right) .
    \end{align}
    \item Once the clients update their model parameters as mentioned in the equation \ref{eq_c1} and \ref{eq_c2}, they send updated model parameters to the federated server. After receiving all updated model parameters $\left\{\boldsymbol{\theta}_{k,t}^\textup{C}, W_{k,t}^\textup{C}\right\}_{k\in[N]}$, the federated server updates the parameters of the global client-side model using federated averaging as: \par
    \begin{equation}
    \boldsymbol{\theta}_{t+1}=\sum_{k \in[N]} \boldsymbol{\theta}_{k,t}^\textup{C} ; \quad W_{t+1}=\sum_{k \in[N]}  W_{k,t}^\textup{C}.
    \end{equation}
\end{enumerate}

For the multi-class classification taking criterion as the cross-entropy loss over the server-side model portion is expressed as:

\begin{equation}
\textup{Cross-entropy loss} = -\frac{1}{S} \sum_{i=1}^{S} \sum_{j=1}^{V} Y_{ij} \log(\frac{\exp(z_{ij})}{\sum_{k=1}^{V} \exp(z_{ik})}),
\end{equation}
where $z_{ij}$ is the server-side model's output (logits) for the smashed data $\boldsymbol{A}_{k,t}[i]$ and class $j$, $S$ denotes the number of samples in the batch, $V$ denotes the number of classes.

\subsection{SFPL with positive labels}

SFLv2 demonstrates significant constraints in managing positive labels, as evidenced by the results presented in Table~\ref{tab:sflv2nonIIDdataset}. To tackle this challenge, we introduce an enhanced version of the SFLv2, which we call SFPL. The algorithm of SFPL is depicted in Algorithm~\ref{alg:cap} and~\ref{alg:cap2}. It comprises four main functions: Client, ClientFedServer, GlobalCollectorFunction, and ServerSideModelFunction. A brief description of each component's functionality, illustrated in Figure \ref{fig:diagram}, is presented in the following:


\begin{algorithm} [t] 
	    \small
		\caption{Splitfed learning with positive labels (SFPL)}
    	   {\textbf{Notations:} (1) At time $t$, $S_t$ is a set of $K$ clients, and $ {\boldsymbol{A}}_{k,t} $ is the smashed data of client $k\in \{1,2,\cdots,K\}$; (2) for any client $k$, (a) ${\boldsymbol{Y}}_k$ and ${\hat{\boldsymbol{Y}}}_k$ are the true and predicted labels, respectively, and (b) $\triangledown\ell_k$ is its gradient of the loss.}

   		\SetKwProg{Fn}{GlobalCollectorFunction$( {\boldsymbol{A}}_{k,t}, {\boldsymbol{Y}}_{k} )$:}{}{}
    \tcc{\scriptsize Runs on Server}	
    \Fn{}{
        Initialize activation and label stack: \\
        $ \textup{ActivationStack} = \phi $ \\
        $ \textup{LabelStack} = \phi $ \\
        \For{each client $ k\in N $, in parallel}{
             Receive $ \boldsymbol{A}_{k,t} $ and $\boldsymbol{Y}_{k}$ from Client $({W}^{\textup{C}}_{k,t})$ \\
             Store ($ \boldsymbol{A}_{k,t} $, ${\boldsymbol{Y}}_k$) in ActivationStack and LabelStack with the client's ID $k$ as its key.}
        Wait until $count(\textup{ActivationStack}) = \alpha N$ \\
        ${\boldsymbol{A'}}_{k,t}$, ${\boldsymbol{Y'}}_k \leftarrow $ Shuffle ($ {\boldsymbol{A}}_{k,t} $, ${\boldsymbol{Y}}_k$) \\
        Send shuffled ActivationStack and LabelStack to ServerSideModelFunction \\
        Wait for the ServerSideModelFunction to execute \\
        Receive ($d{\boldsymbol{A}}_{k,t}$) from  ServerSideModelFunction \\
        De-shuffle $(d{\boldsymbol{A}}_{k,t})$ \\
        Send $d{\boldsymbol{A}}_{k,t}$ back to the respective clients 
    }
    \vspace{5pt}
   	\SetKwProg{Fn}{ServerSideModelFunction($\boldsymbol{A'}_{k,t}$, $\boldsymbol{Y'}_k$)}{}{}
        \tcc{ \scriptsize Runs on Server}
        \Fn{}{
        \eIf {\textup{time instance} t=0}{
        ${W}^{\textup{S}}_{t}$ (global server-side model) is initialized}{
	\textup{Gradient\_collector} = \{\} \\
	\tcc{	\scriptsize ${W}^{\textup{S}}_t$ is continuously updated}  
	    Forward propagation with $ \boldsymbol{A'}_{k,t} $ on $ {W}^{\textup{S}}_t$, compute $ \hat{\boldsymbol{Y'}}_{k} $  \\
	    Loss calculation with $ \boldsymbol{Y'}_{k}$ and  $\hat{\boldsymbol{Y'}}_{k}$ \\    
	    Back-propagation calculate $\triangledown \ell_k ({W}^{\textup{S}}_t; \boldsymbol{A}^{\textup{S}}_t)$\\
	    Collect $ d\boldsymbol{A}_{k,t} := \triangledown\ell_k (\boldsymbol{A}^{\textup{S}}_t; \boldsymbol{W}^{\textup{S}}_t) $ ({ i.e.}, gradient of the $ \boldsymbol{A'}_{k,t} $) in Gradient\_collector.
	
	Send $ d\boldsymbol{A}_{k,t}$ to GlobalCollectorFunction\\
    }}

		\label{alg:cap}
\end{algorithm}

\begin{itemize}
\item \textbf{Client:} Clients initiate with a weight matrix $W_{k,t}^\textup{C}$, perform forward propagation on local data $X_k$, producing smashed data $\boldsymbol{A}_{k,t}$. These, along with true labels $\boldsymbol{Y}_k$, are sent to the GlobalCollectorFunction, returning gradients $d\boldsymbol{A}_{k,t}$. Clients use this for back-propagation, calculating gradients $\triangledown \ell_k({W_{k,t}^\textup{C}})$, updating weights via ${W_{k,t}^\textup{C}}\leftarrow {W_{k,t}^\textup{C}}-\eta \triangledown \ell_k({W_{k,t}^\textup{C}})$, and await ClientFedServer(${W_{k,t}^\textup{C}}$) completion.

\item \textbf{GlobalCollectorFunction:} The global collector function collects activations and true labels from clients, shuffles and sends them to the server-side model function, receives gradient of $\mathbf{A}_{k,t}$, de-shuffles, and sends back to clients.

\item \textbf{ServerSideModelFunction:} The server-side model function receives shuffled activations and labels, initializes global model weights $W^\textup{S}_t$, computes predicted labels $\hat{\boldsymbol{Y}_k}$, evaluates loss and determines gradient $d\mathbf{A}_{k,t}$ to send back to GlobalCollector.

\item \textbf{ClientFedServer:} The function collects client model weights, computes the global model's average weights excluding the batch normalization layer to mitigate its impact, and updates each client's model accordingly.
\end{itemize}
\begin{algorithm} [H] 
	    \small
		
		\caption{Splitfed learning with positive labels (SFPL) client side}

{\textbf{Notations:} (1) For client $k$, the local data is represented by $X_k$, its corresponding true labels are denoted as $\boldsymbol{Y}_{k}$, and neural network layer index by $l$.}

		\SetKwProg{Fn}{Client$  ({W}^{\textup{C}}_{k,t}) $:}{}{} 
            \tcc{	\scriptsize Runs on Client $ k $}	
		\Fn{}{
			Start with ${W}^{\textup{C}}_{k,t}$\\
			Set $ \boldsymbol{A}_{k,t} $ = $ \phi $ \\
				Forward propagation with the local data $\boldsymbol{X}_k$ up to its final layer in $ {W}^{\textup{C}}_{k,t}$ and get the activations $ \boldsymbol{A}_{k,t} $ (smashed data)\\	
				Send $ \boldsymbol{A}_{k,t} $ and $\boldsymbol{Y}_{k}$ to the GlobalCollectorFunction\\
                $d\boldsymbol{A}_{k,t} \leftarrow$ GlobalCollectorFunction($ \boldsymbol{A}_{k,t} $ , $\boldsymbol{Y}_{k}$) \\
				Back-propagation, calculate gradients $ \triangledown  \ell_k ({W}^{\textup{C}}_{k,t})$ with $d\boldsymbol{A}_{k,t}$ \\
                Update ${W}^{\textup{C}}_{k,t}\leftarrow {W}^{\textup{C}}_{k,t}-\eta \triangledown  \ell_k ({W}^{\textup{C}}_{k,t})$	\\
                Calls and wait for the completion of ClientFedServer(${W}^{\textup{C}}_{k,t}$)\\
			
			}
\vspace{5pt}
              \SetKwProg{Fn}{ClientFedServer$({W}^{\textup{C}}_{k,t})$:}{}{}
    \tcc{	\scriptsize Runs on Server}
    \Fn{}{
        Initialize ${W}^{\textup{avg}}_{t}$ \\
        \For{each client $ k\in N $, in parallel}{
            Receive updated model weights ${W}^{\textup{C}}_{k,t}$ from client $k$ \\
        }
    
        Compute the average model weights: \\
        \For{each client $k$ and each layer l}{
        \If {\textup{layer \emph{l} is not BatchNorm}}{
            ${W}^{\textup{avg}(l)}_{t} = \frac{1}{K}\sum_{k=1}^{K} {W}^{\textup{C}(l)}_{k,t}$ 
        }
        }
                
        \For{each client $ k\in N $, in parallel}{
            Update ${W}^{\textup{C}}_{k,t}\leftarrow 
            {W}^{\textup{avg}}_{t}$
        }
        }
        
		\label{alg:cap2}
	\end{algorithm}

\begin{table*}[!htb]
\begin{center}
\caption{Impact of IID and non-IID distribution with only positive labels on SFLv2 on CIFAR-10 dataset.}
\label{tab:sflv2nonIIDdataset} 
\begin{tabular}{ccccccccc}
\hline
\\[-6pt]
Architecture & SD and VAR & Training IID & Testing IID & Precision@1 & Recall & F1Score & Accuracy & Loss \\
\hline
\\[-6pt]
 R32 & RMSD & $\checkmark$ & $\checkmark$ & 0.5098 & 0.5049 & 0.4963 & 50.49 & 1.346 \\
& RMSD & $\times$ & $\times$ & 0.01 & 0.1 & 0.01818 & 10 & 8.89 \\
& RMSD & $\times$ & $\checkmark$ & 0.2082 & 0.1327 & 0.06871 & 13.27 & 2.439 \\

R8 & RMSD & $\checkmark$ & $\checkmark$ & 0.7457 & 0.7402 & 0.739 & 74.02 & 0.7616 \\
& RMSD & $\times$ & $\times$ & 0.01 & 0.1 & 0.0181 & 10 & 7.481 \\
& RMSD & $\times$ & $\checkmark$ & 0.03209 & 0.1091 & 0.03354 & 10.91 & 4.815 \\
\hline
\\[-6pt]
\end{tabular}
\end{center}
\end{table*}

\begin{table*}[!htb]
\centering
\caption{Total cost analysis of the three DCML approaches for one global epoch.}
\label{tab:comm cost} 
\begin{tabular}{cccc}
\hline
\\[-6pt]
Method & Comms. per client & Total comms. per client  & Total model training time   \\
\hline
\\[-6pt]
FL & $2|\mathbf{W}|$ & $2 N|\mathbf{W}|$ & $T+2 \frac{|\mathbf{W}|}{R}+T_{\text {fedavg }}$ \\
SFLv2 & $\left(\frac{2 \mathcal{X}}{N}\right) q+2 \beta|\mathbf{W}|$ & $2 \mathcal{X} q+2 \beta N|\mathbf{W}|$ & $T+2 \frac{\mathcal{X} q}{N R}+2 \frac{\beta|\mathbf{W}|}{R}+\frac{T_{\text {fedavg }}}{2}$ \\
SFPL & $\left(\frac{2 \mathcal{X}}{N}\right) q+2 \beta|\mathbf{W}|$ & $2 \mathcal{X} q+2 \beta N|\mathbf{W}|$ & $T+2 \frac{\mathcal{X} q}{N R}+2 \frac{\beta|\mathbf{W}|}{R}+\frac{T_{\text {fedavg }}}{2}$  \\ 
\\
\hline
\\[-6pt]
\end{tabular}
\end{table*}

\section{Failure analysis of splitfed learning for positive labels}
\label{sec:failuresflv2}
This section presents an analysis of SFLv2's failures. This study is applicable to other SFL variants, such as SFLv1. SFLv2 suffers due to the following when engaged in a learning scenario with only positively labeled data (see Table~\ref{tab:sflv2nonIIDdataset}). 

\subsection{Failure caused by catastrophic forgetting}
Catastrophic interference is a phenomenon observed in machine learning where a model loses previously acquired knowledge upon being fine-tuned on a new task. In SFLv2, catastrophic interference arises on the server side when client-side models train on single-class data. The server-side model encounters clients' smashed data in random, sequentially ordered batches during each epoch, treating each batch as a learning task. SFLv2 aims to minimize cumulative loss across tasks. In a scenario with $N$ clients, each task $q \in \{1, 2, \dots, Q_n\}$, where $Q_n$ represents the task of learning labels for the $n$-th client, can be considered sequential tasks for the server-side model. With $L_q(\theta_q)$ representing the loss on task $q$, the optimal total loss can be expressed as:
\begin{equation}
L(\theta_1, \theta_2, \dots, \theta_{n}) = \sum_{q=1}^{Q_n} L_q(\theta_q).
\end{equation} 

\begin{figure}[t]
  \centering
  \includegraphics[width=3.2in]{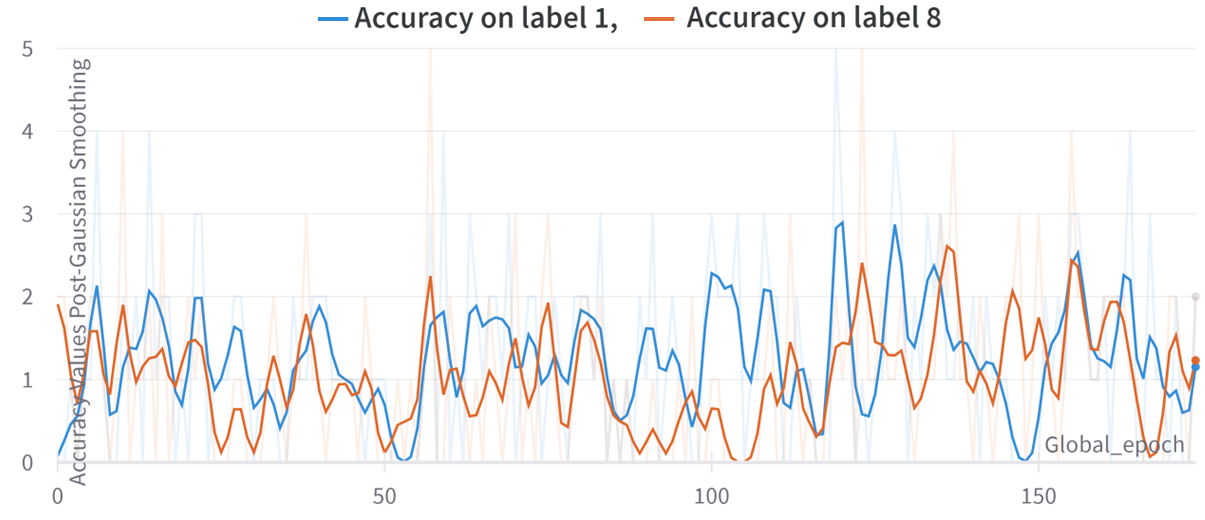}
  \caption{Accuracy Values Post-Gaussian Smoothing recorded for labels 1 and 8 for SFLv2.}
  \label{fig:diagram3}
\end{figure}
As the server-side model trains sequentially with batches from each client in an epoch, it exhibits higher accuracy for the class associated with the last visited client and lower accuracy for the class of the first visited client. 
The graphical representation depicted in Figure~\ref{fig:diagram3} presents a pattern of accuracy dynamics for labels 1 and 8 during the training process of the CIFAR-100 dataset~\cite{krizhevsky2009learning} using the SFLv2 model. Following a certain number of epochs, a steep decrease in accuracy for label 8 is noticeable post the 25th epoch, coinciding with a significant increase in accuracy for label 1. It is noteworthy that in circumstances where a marked boost in accuracy for a specific label is detected, the final iteration of the batch SFLv2 model is primarily linked with the client associated with that particular label, in this instance, label 1. This pattern undergoes a shift at the 148th epoch, where a sharp ascent in accuracy for label 8 is observed, accompanied by a decline in accuracy for label 1. This shift is again correlated with the last label being trained upon, which in this case, is label 8.

\subsection{Failure caused by weight divergence} 
    Weight divergence in the context of non-IID refers to the discrepancy in the model weights of SFLv2 when compared to those of standard stochastic gradient descent (SGD). Weight divergence in SFLv2 learning refers to the discrepancy between aggregated client-side model parameters under two different data distribution scenarios: IID data and non-IID data on the client side. In a study~\cite{zhao2018federated}, the authors demonstrated that aggregating FL models in a non-IID setting results in model weight divergence when compared to FL models in an IID setting. Weight divergence statistics were utilized to measure this divergence.\par
    For SFLv2, weight divergence statistics can be defined with certain assumptions. Suppose the client-side global model, after an epoch $e$ with an IID distribution, is $\boldsymbol{w}^{\textup{
    SGD}}$, and with a non-IID distribution after the same number of epochs, is $\boldsymbol{w}^{\textup{FedAvg}}$. The weight divergence statistics can be defined as:

    \begin{equation}
        \textup{weightDivergence}=\frac{\left|\boldsymbol{w}^{\text{FedAvg}}-\boldsymbol{w}^{\text{SGD}}\right|}{\left|\boldsymbol{w}^{\text{SGD}}\right|}.
    \end{equation}

    The theoretical and empirical evaluations showed weight divergence for federated learning (FL) with both IID and non-IID data~\cite{zhao2018federated}. This remains applicable to SFLv2, where client-side model aggregation occurs under IID and non-IID data scenarios on the client side.
    
 \subsection{Failure caused by batch normalization layer at client-side model portion} \par
    In SFLv2, the batch normalization layer parameters, including the mean and variance, are typically aggregated as part of the client-side model portion.  
    However, studies in~\cite{li2021fedbn} assert that in a non-IID FL framework, model parameters are contingent upon each client's local dataset and influenced by its data distribution. Designating the $k$-th client's activation mean and variance as $\mu_k$ and $\sigma_k^2$ respectively, the batch normalization layer normalizes these using the following equation:
    \begin{equation}
    \hat{x}_i = \frac{x_i - \mu_k}{\sqrt{\sigma_k^2 + \epsilon}},
    \end{equation}
    where $x_i$ is the activation of the $i$-th neuron, $\epsilon$ is a small constant added to avoid division by zero, and $\hat{x}_i$ is the normalized activation.
    As each client's local data distribution may vary, their mean and variance estimates can also differ. Averaging the means and variances of all clients may not accurately capture the distribution of activations for each individual client. In the study by Li et al. \cite{li2021fedbn}, it has been demonstrated both theoretically and empirically that in federated learning (FL), the global model converges more quickly when batch normalization (BN) layers are not aggregated. This finding is extended to SFLv2, where client-side model weights are aggregated at the end of each epoch.

\begin{table*}[!tbh]
\centering
\caption{Dataset}
\label{tab:dataset} 
\begin{tabular}{ccccccc}
\hline
\\[-6pt]
Dataset & \#Features & \#Labels Or \#Clients & \#TrainPoints & \#TestPoints  & Training Instances/ class (or client) \\
\hline
\\[-6pt]
CIFAR10 & 1024 (32x32) & 10 & 50000 & 10000  & 5000 \\ 
CIFAR100 & 1024 (32x32) & 100 & 50000 & 10000  & 500 \\ 
\\
\hline
\\[-6pt]
\end{tabular}
\end{table*}

\begin{table*}[!tbh]
\centering
\caption{Flops allowed per resource-constrained IoT client}
\label{tab:flops}
\begin{tabular}{ccccccc}
\hline
\\[-6pt]
Architecture & Dataset & Client flops/datapoint & Client Params & Server Flops & Server Params  &  Kernel Size \\
\hline
\\[-6pt]
R8 & CIFAR-10 & 475.136K & 464 & 2.47014M & 74.826K  &  $(7 \times 7), (3 \times 3)$  \\
R32 & CIFAR-10 & 475.136K & 464 & 14.6691M & 463.69K & $(7 \times 7), (3 \times 3)$  \\
R32 & CIFAR-100 & 475.136K & 464 & 14.7612M & 469.54K & $(7 \times 7), (3 \times 3)$  \\
R56 & CIFAR-100 & 475.136K & 464 & 26.9601M & 858.404K & $(7 \times 7), (3 \times 3)$  \\
\\
\hline
\\[-6pt]
\end{tabular}
\end{table*}

\begin{table*}[!htb]
\caption{Improvement from base SFLv2 to SFPL and comparison with FedAws}
\label{tab:exp1}
\centering
\begin{tabular}{ccccccccccc}
\\
\hline
\\[-6pt]
Dataset & Architecture & Technique & Training IID & Testing IID & SD and VAR & Precision@1 & Recall & F1Score & Accuracy \\
\hline
\\[-6pt]
CIFAR-100 & R56 & SFPL & $\times$ & $\times$ & CMSD & 0.7331 & 0.7216 & 0.7215 & 72.16\\ 
 &  & SFPL & $\times$ & $\times$ & RMSD & 0.6623 & 0.6313 & 0.628 & 63.13  \\ 
 &  & SFPL & $\times$ & $\checkmark$ & CMSD & 0.6319 & 0.598 & 0.5915 & 59.8  \\
 &  & SFLv2 & $\times$ & $\times$ & RMSD & 0.0047 & 0.014 & 0.0036 & 1.4  \\
 &  & FedAws & $\times$ & - & - & 0.696 & - & - & -  \\
CIFAR-100 & R32 & SFPL & $\times$ & $\times$ & CMSD & 0.6955 & 0.6677 & 0.666 & 66.77  \\
 &  & SFPL & $\times$ & $\times$ & RMSD & 0.6265 & 0.5926 & 0.5811 & 59.26 \\ 
 &  & SFPL & $\times$ & $\checkmark$ & CMSD & 0.608 & 0.5542 & 0.5475 & 55.42  \\ 
 &  & SFLv2 & $\times$ & $\times$ & RMSD & 0.0213 & 0.0205 & 0.008156 & 2.05  \\ 
 &  & FedAws & $\times$ & - & - & 0.679 & - & - & - \\
CIFAR-10 & R32 & SFPL & $\times$ & $\times$ & CMSD & 0.9232 & 0.9233 & 0.923 & 92.33 \\
 &  & SFPL & $\times$ & $\times$ & RMSD & 0.8905 & 0.8859 & 0.8858 & 88.59  \\ 
 &  & SFPL & $\times$ & $\checkmark$ & CMSD & 0.8909 & 0.8883 & 0.8884 & 88.83  \\ 
 &  & SFLv2 & $\times$ & $\times$ & RMSD & 0.01 & 0.1 & 0.01818 & 10  \\ 
 &  & FedAws & $\times$ & - & - & 0.924 & - & - & - \\
CIFAR-10 & R8 & SFPL & $\times$ & $\times$ & CMSD & 0.8537 & 0.8515 & 0.8508 & 85.15  \\
& & SFPL & $\times$ & $\times$ & RMSD & 0.8049 & 0.7989 & 0.7987 & 79.89  \\
& & SFPL & $\times$ & $\checkmark$ & CMSD & 0.7847 & 0.7743 & 0.7739 & 77.43  \\
& & SFLv2 & $\times$ & $\times$ & RMSD & 0.01 & 0.1 & 0.0181 & 10  \\
& & FedAws & $\times$ & - & - & 0.863 & - & - & -  \\
\\
\hline
\\[-6pt]
\end{tabular}
\end{table*}

\section{Communication size and model training time equations}
This section presents an analysis of communication cost and model training time for SFPL, SFLv2, and FL. Let $N$ represent the number of clients, $\mathcal{X}$ denote the total size of the dataset, $q$ be the size of the smashed layer, $R$ be the communication rate, and $T$ be the time required for one forward and backward propagation on the full model using a dataset of size $\mathcal{X}$ (for any architecture). Let $T_{\text {fedavg}}$ denote the time required for full model aggregation (with $\frac{T_{\text {fedavg}}}{2}$ being the aggregation time for client-side model aggregation only), $|\mathbf{W}|$ represent the size of the full model, and $\beta$ be the fraction of the full model's size available in a client for SFPL/SFL, \emph{i.e.}, $\left|\mathbf{W}^{\mathrm{C}}\right|=\beta|\mathbf{W}|$. The term $2 \beta|\mathrm{W}|$ indicates the communication cost per client due to the download and upload of the client-side model updates before and after training, respectively. Despite the addition of a global collector function, the latency incurred can be considered negligible due to its location on the same server as the server-side model function. The results are presented in Table \ref{tab:comm cost}. As observed in the table, the total training time cost increases in the order of FL $<$ SFLv2 $=$ SFPL as the number of clients ($N$) increases.

\section{Experiments}
\label{exp}
This section delineates the empirical study's dataset, model architecture, and initialization strategy, followed by the comparative experimental results of the proposed SFPL framework, SFLv2, and FedAws. Further examinations to substantiate the proposed framework's efficacy under diverse conditions are also presented. Our experiments employ the CIFAR-10 and CIFAR-100 image datasets \cite{krizhevsky2009learning}, each containing 50,000 training and 10,000 test images of 3072 dimensions $(32 \times 32)$. To mitigate overfitting, random horizontal flipping, rotation, normalization, and cropping were applied. Following the data partitioning strategy in Section~\ref{sec:ps}, each client corresponds to a single class; thus, for CIFAR-10, 10 clients are considered, while for CIFAR-100, 100 clients are selected. As outlined in Table~\ref{tab:dataset}, we utilized three primary model architectures, R8, R32, and R56 \cite{he2016deep}, because of the discrete "blocks" structure which facilitates model partitioning. \par 
Given the study's emphasis on resource-constrained IoT clients, these are identified as devices limited to $475.136$K floating-point operations per second (Flops), a computational capacity compared to the requirements of the initial block in the Resnet model architecture. Server flops and parameters escalated with architectural complexity (Table \ref{tab:flops}). The experiment's initialization parameters comprised a learning rate of $1 \times 10^{-1}$, mini-batch size of $4$, gamma parameter of $2 \times 10^{-2}$, weight decay parameter of $5 \times 10^{-4}$, momentum parameter of $9 \times 10^{-1}$, and the MultiStepLR function for learning rate decay. The communication round was set at 175, with milestones at the $60^{th}$, $120^{th}$, and $160^{th}$ epochs, and kernel sizes of 7 $\times$ 7 and 3 $\times$ 3, as encapsulated in Table~\ref{tab:flops}.\par
To emulate real-world conditions and effectively illustrate SFPL's behavior, we have chosen three testing scenarios for our evaluation, as used in Tables ~\ref{tab:exp1},~\ref{tab:IID_impact},~\ref{tab:nonIID},~\ref{tab:nonIIDtrainingIID testing}: 
\begin{itemize}
    \item Training IID (represented as: Training IID $\checkmark$), Testing IID (represented as: Testing IID $\checkmark$): This scenario involves training and testing an AI model under identical distributional conditions using IID data. The model is trained on a representative real-world data distribution and subsequently tested on a separate IID test set. This provides a performance benchmark under similar distributional conditions.
    \item Training extreme non-IID (represented as: Training IID $\times$), Testing IID (represented as: Testing IID $\checkmark$): Here, the model is trained on extreme non-IID data, where each client retains its data without sharing. Post-training, the model is evaluated on an IID test set. This scenario assesses the model's performance when trained on extreme non-IID data but tested on IID data, reflecting practical situations with uneven data distribution across clients.
    \item Training extreme non-IID (represented as: Training IID $\times$), Testing extreme non-IID (represented as: Testing IID $\times$): This scenario mirrors the previous one, but with testing also conducted on extreme non-IID data. For example, in user identification tasks like speaker recognition, only test instances corresponding to the specific speaker are used. This evaluates the model's performance under persistent extreme non-IID conditions during both training and testing.
\end{itemize}

\begin{table*}[!htb]
\centering
\caption{Impact of IID training and testing on SFPL when batch normalization is set to current mean and variance instead of running mean and variance}
\label{tab:IID_impact}
\begin{tabular}{ccccccccc}
\hline
\\[-6pt]
Dataset & Architecture & SD and VAR & Training IID & Testing IID & Precision@1 & Recall & F1Score & Accuracy \\
\hline
\\[-6pt]
CIFAR-100 & R56 & RMSD & $\checkmark$ & $\checkmark$ & 0.7062 & 0.703 & 0.7035 & 70.3  \\
CIFAR-100 & R56 & CMSD & $\checkmark$ & $\checkmark$ & 0.6965 & 0.6919 & 0.6915 & 69.19 \\
CIFAR-100 & R32 & RMSD & $\checkmark$ & $\checkmark$ & 0.6868 & 0.6845 & 0.6843 & 68.45 \\
CIFAR-100 & R32 & CMSD & $\checkmark$ & $\checkmark$ & 0.6748 & 0.6698 & 0.6694 & 66.98  \\
CIFAR-10 & R32 & RMSD & $\checkmark$ & $\checkmark$ & 0.9203 & 0.9202 & 0.9201 & 92.02 \\
CIFAR-10 & R32 & CMSD & $\checkmark$ & $\checkmark$ & 0.9175 & 0.9174 & 0.9173 & 91.74  \\
CIFAR-10 & R8 & RMSD & $\checkmark$ & $\checkmark$ & 0.8557 & 0.8553 & 0.8552 & 85.53 \\
CIFAR-10 & R8 & CMSD & $\checkmark$ & $\checkmark$ & 0.8385 & 0.839 & 0.8373 & 83.9\\
\\
\hline
\\[-6pt]
\end{tabular}
\end{table*}

\begin{table*}[!htb]
\caption{Impact of non-IID training and IID testing on SFPL when batch normalization is set to current mean and variance instead of running mean and variance}
\label{tab:nonIIDtrainingIID testing}
\centering
\begin{tabular}{cccccccccc}
\hline
\\[-6pt]
Dataset & Architecture & SD and VAR & Training IID & Testing IID & Precision@1 & Recall & F1Score & Accuracy & Loss \\
\hline
\\[-6pt]
CIFAR-100 & R56 & RMSD & $\times$ & $\checkmark$ & 0.6637 & 0.6318 & 0.6225 & 63.18 & 1.668 \\
CIFAR-100 & R56 & CMSD & $\times$ & $\checkmark$ & 0.6319 & 0.598 & 0.5915 & 59.8 & 1.825 \\
CIFAR-100 & R32 & RMSD & $\times$ & $\checkmark$ & 0.6233 & 0.5863 & 0.5749 & 58.63 & 1.747 \\
CIFAR-100 & R32 & CMSD & $\times$ & $\checkmark$ & 0.608 & 0.5542 & 0.5475 & 55.42 & 1.902 \\
CIFAR-10 & R32 & RMSD & $\times$ & $\checkmark$ & 0.898 & 0.8963 & 0.8964 & 89.63 & 0.3686 \\
CIFAR-10 & R32 & CMSD & $\times$ & $\checkmark$ & 0.8909 & 0.8883 & 0.8884 & 88.83 & 0.39 \\
CIFAR-10 & R8 & RMSD & $\times$ & $\checkmark$ & 0.8208 & 0.819 & 0.8182 & 81.9 & 0.5567 \\
CIFAR-10 & R8 & CMSD & $\times$ & $\checkmark$ & 0.7847 & 0.7743 & 0.7739 & 77.43 & 0.6922 \\
\\
\hline
\\[-6pt]
\end{tabular}
\end{table*}
\begin{table*}[!htb]
\caption{Impact of non-IID training and non-IID testing on SFPL when batch normalization is set to current mean and variance instead of running mean and variance}
\label{tab:nonIID}
\centering
\begin{tabular}{cccccccccc}
\hline
\\[-6pt]
Dataset & Architecture & SD and VAR & Training IID & Testing IID & Precision@1 & Recall & F1Score & Accuracy & Loss \\
\hline
\\[-6pt]
CIFAR-100 & R56 & RMSD & $\times$ & $\times$ & 0.6623 & 0.6313 & 0.628 & 63.13 & 1.604 \\
CIFAR-100 & R56 & CMSD & $\times$ & $\times$ & 0.7331 & 0.7216 & 0.7215 & 72.16 & 1.133 \\
CIFAR-100 & R32 & RMSD & $\times$ & $\times$ & 0.6265 & 0.5926 & 0.5811 & 59.26 & 1.742 \\
CIFAR-100 & R32 & CMSD & $\times$ & $\times$ & 0.6955 & 0.6677 & 0.666 & 66.77 & 1.25 \\
CIFAR-10 & R32 & RMSD & $\times$ & $\times$ & 0.8905 & 0.8859 & 0.8858 & 88.59 & 0.4166 \\
CIFAR-10 & R32 & CMSD & $\times$ & $\times$ & 0.9232 & 0.9233 & 0.923 & 92.33 & 0.2714 \\
CIFAR-10 & R8 & RMSD & $\times$ & $\times$ & 0.8049 & 0.7989 & 0.7987 & 79.89 & 0.6753 \\
CIFAR-10 & R8 & CMSD & $\times$ & $\times$ & 0.8537 & 0.8515 & 0.8508 & 85.15 & 0.4419 \\
\\
\hline
\\[-6pt]
\end{tabular}
\end{table*}

\subsection{Performance of DL model training with only positive labels on SFPL, SFLv2, and FedAws}

In this section, we assess the performance of SFPL, SFLv2, and FedAws. It is important to acknowledge that the FedAws framework may not be optimally suited for resource-constrained IoT devices with computational limitations capped at $475.136$ K Flops. Nevertheless, we incorporate FedAws in our comparative analysis, as it shares a comparable learning setting that exclusively relies on positive labels, thereby providing valuable insights and context for the evaluation of SFPL and SFLv2. For experiments with SFPL and SFLv2 frameworks, the DL model under consideration was partitioned at the first layer, with the initial layer assigned to the client side and the rest of the model allocated to the server side. Finally, we assess the experiments in terms of precision@1 and F1-score on the CIFAR-10 and CIFAR-100 test datasets. Additionally, recall and accuracy metrics are included in our evaluation table to offer complementary perspectives that contribute to a comprehensive assessment of the model's performance.

Additionally, we examine the impact of using the current mean and variance (CMSD) strategy, where the batch normalization layer for the client-side model portion is not aggregated during the aggregation step. Instead, the mean and variance of the batch under test are utilized during testing. In contrast, the running means and standard deviation (RMSD) strategy aggregates the batch normalization layer for the client-side model portion during the aggregation step, using the learned running mean and variance for the batch normalization layer during testing. It is important to note that due to the unavailability of the FedAws code and the inability to reproduce results, we have taken the FedAws results from its ICLR research paper~\cite{lin2022federated}.

The empirical findings presented in Table~\ref{tab:exp1} demonstrate the efficacy of SFPL in mitigating the constraints inherent in the SFLv2 learning framework while training DL models solely utilizing positive labels. In the CIFAR-100 dataset, SFLv2 reported relatively low accuracies, approximately 1.4\% for the R56 architecture and 2.05\% for the R32 architecture, indicating a failure to learn effectively. Conversely, the SFPL framework significantly enhanced performance, achieving 72.16\% and 66.77\% accuracy for the R56 and R32 architectures respectively. A similar trend was observed in the CIFAR-10 dataset, with the R32 and R8 architectures' accuracies stagnating at 10\% under SFLv2, while SFPL significantly boosted them to 92.33\% and 85.15\%, respectively.

Table~\ref{tab:exp1} also presents additional evaluation metrics, such as Precision@1, recall, and F1 score, to corroborate that the DL models effectively learned all classes when employing the SFPL framework. Furthermore, the results indicate that the CMSD setup yielded superior outcomes compared to the RMSD setup during the testing phase, specifically when a single-class batch was utilized for model evaluation. Conversely, SFPL's performance declined when an IID batch was employed for testing.

Lastly, SFPL surpassed FedAws' Precision@1 scores for the CIFAR-100 R-56 and R-32 architectures by 3.71\% and 1.65\%, respectively. Nevertheless, the performance remained comparable for the CIFAR-10 R-32 and R-8 architectures.

\subsection{More study}
In this section, we present an extended study to analyze the performance of the SFPL framework in various training and testing scenarios. Specifically, we investigate three sub-settings of the SFPL framework, as discussed in section~\ref{exp}.
For each sub-setting, we evaluate the performance of the SFPL framework while introducing both CMSD and RMSD setup on the client side. This study aims to gain insights into the effectiveness and limitations of the SFPL framework across these different scenarios.
\begin{enumerate}
\item \textbf{SFPL: IID dataset at training and inference phase}\par
Based on the results presented in Table~\ref{tab:IID_impact}, it can be inferred that the RMSD setup outperforms the CMSD setup for all combinations of model architectures and datasets considered. These findings suggest that, within the context of the SFPL framework, performing BN layer aggregation during the client-side model aggregation process contributes to improved performance during the inference phase.
\item \textbf{SFPL: non-IID dataset at training and IID test dataset at inference phase}\par
The results in Table~\ref{tab:nonIIDtrainingIID testing} reveal that the RMSD setup consistently surpasses the CMSD setup across all examined model architectures and dataset combinations. This indicates that within the SFPL framework, employing BN layer aggregation during client-side model aggregation for non-IID training and IID inference scenarios leads to enhanced performance during inference.
\item \textbf{SFPL: non-IID dataset at training and inference phase}\par
The results in Table~\ref{tab:nonIID} demonstrate that the CMSD setup consistently surpasses the RMSD setup across all model architectures and dataset combinations, a considerable performance difference. This can be attributed to the utilization of non-IID testing datasets, enhancing the accuracy of the BN layer's statistics for the client-side model in the CMSD setup through the test batch's current mean and variance. Thus, in the SFPL framework for non-IID training and testing, aggregating the batch normalization layer during the training aggregation phase could potentially detriment the overall performance.
\end{enumerate}

\section{Conclusion}
In this paper, we proposed an algorithm called splitfed learning with positive labels (SFPL) for developing better classifiers in a DCML framework with resource-constrained IoT environments, where all clients have only positive labels. SFPL integrated a global collector on the server side along with the attenuation of the batch normalization layer on the client-side model portions. We demonstrated empirically that SFPL outperformed SFLv2 for extreme non-IID scenarios.
Furthermore, we conducted an in-depth study demonstrating that altering the aggregation strategy during client-side model weight aggregation in SFPL produced high-quality models for both IID and non-IID data. Thus, SFPL offered an effective solution for mitigating challenges stemming from resource-constrained IoT devices when exposed to exclusively positive labels. This paper serves as an initial step toward addressing these challenges. Future research could explore more experiments involving diverse models and other datasets with more classes.



\ifCLASSOPTIONcaptionsoff
  \newpage
\fi



%


\begin{thebibliography}{319}

\ifx \showCODEN    \undefined \def \showCODEN     #1{\unskip}     \fi
\ifx \showDOI      \undefined \def \showDOI       #1{#1}\fi
\ifx \showISBNx    \undefined \def \showISBNx     #1{\unskip}     \fi
\ifx \showISBNxiii \undefined \def \showISBNxiii  #1{\unskip}     \fi
\ifx \showISSN     \undefined \def \showISSN      #1{\unskip}     \fi
\ifx \showLCCN     \undefined \def \showLCCN      #1{\unskip}     \fi
\ifx \shownote     \undefined \def \shownote      #1{#1}          \fi
\ifx \showarticletitle \undefined \def \showarticletitle #1{#1}   \fi
\ifx \showURL      \undefined \def \showURL       {\relax}        \fi
\providecommand\bibfield[2]{#2}
\providecommand\bibinfo[2]{#2}
\providecommand\natexlab[1]{#1}
\providecommand\showeprint[2][]{arXiv:#2}

\bibitem{konevcny2016federated}
J.~Kone{\v{c}}n{\`y}, H.~B. McMahan, F.~X. Yu, P.~Richt{\'a}rik, A.~T. Suresh,
  and D.~Bacon, ``Federated learning: Strategies for improving communication
  efficiency,'' \emph{arXiv preprint arXiv:1610.05492}, 2016.

\bibitem{joshi2022performance}
P.~Joshi, C.~Thapa, S.~Camtepe, M.~Hasanuzzaman, T.~Scully, and H.~Afli,
  ``Performance and information leakage in splitfed learning and multi-head
  split learning in healthcare data and beyond,'' \emph{Methods and Protocols},
  vol.~5, no.~4, p.~60, 2022.

\bibitem{thapa2022splitfed}
C.~Thapa, P.~C.~M. Arachchige, S.~Camtepe, and L.~Sun, ``Splitfed: When
  federated learning meets split learning,'' in \emph{Proceedings of the AAAI
  Conference on Artificial Intelligence}, vol.~36, no.~8, 2022, pp. 8485--8493.

\bibitem{gao2021evaluation}
Y.~Gao, M.~Kim, C.~Thapa, A.~Abuadbba, Z.~Zhang, S.~Camtepe, H.~Kim, and
  S.~Nepal, ``Evaluation and optimization of distributed machine learning
  techniques for internet of things,'' \emph{IEEE Transactions on Computers},
  vol.~71, no.~10, pp. 2538--2552, 2021.

\bibitem{lin2022federated}
X.~Lin, H.~Chen, Y.~Xu, C.~Xu, X.~Gui, Y.~Deng, and Y.~Wang, ``Federated
  learning with positive and unlabeled data,'' in \emph{International
  Conference on Machine Learning}.\hskip 1em plus 0.5em minus 0.4em\relax PMLR,
  2022, pp. 13\,344--13\,355.

\bibitem{mcmahan2017communication}
B.~McMahan, E.~Moore, D.~Ramage, S.~Hampson, and B.~A. y~Arcas,
  ``Communication-efficient learning of deep networks from decentralized
  data,'' in \emph{Artificial intelligence and statistics}.\hskip 1em plus
  0.5em minus 0.4em\relax PMLR, 2017, pp. 1273--1282.

\bibitem{vepakomma2018no}
P.~Vepakomma, T.~Swedish, R.~Raskar, O.~Gupta, and A.~Dubey, ``No peek: A
  survey of private distributed deep learning,'' \emph{arXiv preprint
  arXiv:1812.03288}, 2018.

\bibitem{krizhevsky2009learning}
A.~Krizhevsky and G.~Hinton, ``Learning multiple layers of features from tiny
  images,'' \emph{University of Toronto}, vol.~1, no.~4, p.~7, 2009.

\bibitem{zhao2018federated}
Y.~Zhao, M.~Li, L.~Lai, N.~Suda, D.~Civin, and V.~Chandra, ``Federated learning
  with non-iid data,'' \emph{arXiv preprint arXiv:1806.00582}, 2018.

\bibitem{li2021fedbn}
X.~Li, M.~Jiang, X.~Zhang, M.~Kamp, and Q.~Dou, ``Fedbn: Federated learning on
  non-iid features via local batch normalization,'' \emph{arXiv e-prints}, pp.
  arXiv--2102, 2021.

\bibitem{he2016deep}
K.~He, X.~Zhang, S.~Ren, and J.~Sun, ``Deep residual learning for image
  recognition,'' in \emph{Proceedings of the IEEE conference on computer vision
  and pattern recognition}, 2016, pp. 770--778.

\end{thebibliography}
\bibliographystyle{ACM-Reference-Format}  




\end{document}